\documentclass{article}
\usepackage{amsmath,amssymb,amscd,amsthm,mathtools,amsfonts,latexsym}
\usepackage[dvipdfmx]{graphicx}
\usepackage[dvipdfmx]{color}
\usepackage{ascmac,fancybox}
\usepackage{latexsym}
\usepackage{here}
\usepackage{bm}
\usepackage{float}
\usepackage{color}
\usepackage{algorithm}
\usepackage{algpseudocode}
\usepackage{comment}
\usepackage{multirow}
\usepackage{multicol}

\usepackage{amsthm}
\theoremstyle{definition}
\newtheorem{dfn}{Definition}

\usepackage{threeparttable}
\usepackage{wrapfig}
\usepackage{enumitem}
\usepackage{xcolor}
\usepackage{booktabs}
\usepackage{cite}

\begin{document}

\title{An Algorithmic Framework for Constructing Multiple Decision Trees by Evaluating Their Combination Performance Throughout the Construction Process}
\author{Keito Tajima, Naoki Ichijo, Yuta Nakahara, and Toshiyasu Matsushima}
\date{}
\maketitle

\begin{abstract}
  Predictions using a combination of decision trees are known to be effective in machine learning. Typical ideas for constructing a combination of decision trees for prediction are bagging and boosting. Bagging independently constructs decision trees without evaluating their combination performance and averages them afterward. Boosting constructs decision trees sequentially, only evaluating a combination performance of a new decision tree and the fixed past decision trees at each step. Therefore, neither method directly constructs nor evaluates a combination of decision trees for the final prediction. When the final prediction is based on a combination of decision trees, it is natural to evaluate the appropriateness of the combination when constructing them. In this study, we propose a new algorithmic framework that constructs decision trees simultaneously and evaluates their combination performance throughout the construction process. Our framework repeats two procedures. In the first procedure, we construct new candidates of combinations of decision trees to find a proper combination of decision trees. In the second procedure, we evaluate each combination performance of decision trees under some criteria and select a better combination. To confirm the performance of the proposed framework, we perform experiments on synthetic and benchmark data.
\end{abstract}

\section{Introduction}
\label{section1}
The prediction problem is of great importance in the field of machine learning, and decision tree algorithms are frequently employed to address this problem.
When predicting with a single decision tree, the decision tree is first constructed from the training data under some criteria. Then, the prediction is obtained by inputting a new explanatory variable into the constructed decision tree. CART~\cite{CART}, ID3~\cite{ID3}, and C4.5~\cite{C4.5} are well-known methods for constructing a single decision tree.
On the other hand, constructing $B$ decision trees and weighting the predictions from each tree is known to be effective, where $B$ denotes the number of decision trees for prediction.  We call this prediction ``tree-combined prediction.'' 

Bagging and boosting are the main approaches for ``tree-combined prediction.''
When considering ``tree-combined prediction'' as the final prediction, it is natural to evaluate the performance of ``tree-combined prediction'' throughout the construction process. However, both bagging and boosting construct $B$ decision trees without evaluating the performance of ``tree-combined prediction'' throughout the construction process.
In bagging, $B$ decision trees are constructed independently and simultaneously under some criteria. The final prediction is the average of the predictions from each decision tree. 
Random Forests (RF)~\cite{breiman2001random} and  Extremely Randomized Trees (ET)~\cite{ET} are famous methods that use the idea of bagging.
In boosting, $B$ decision trees are constructed by repeatedly adding the decision trees that fit the residuals.
The final prediction is the weighting of the predictions from each decision tree. Algorithms that use boosting include AdaBoost~\cite{Boosting}, gradient boosting decision tree (GBDT)~\cite{GBDT}, XGBoost~\cite{XGBoost}, and LightGBM~\cite{LightGBM}.

In this study, we propose a new algorithmic framework for evaluating the performance of ``tree-combined prediction'' when constructing $B$ decision trees. 
In other words, we construct a combination of $B$ decision trees directly evaluating the performance of the final prediction. In our framework, the operations named ``grow'' and ``select'' are iterated to construct $B$ decision trees. The final prediction is the weighted prediction of the predictions from each tree.  
In ``grow,'' we grow $B$ decision trees to create a new set of decision trees.
Each decision tree creates new $B_{\text{keep}}$ decision trees by splitting one leaf node.
Therefore, a new set of $B\cdot B_{\text{keep}}$ decision trees is obtained.
In ``select,'' we consider many combinations of $B$ decision trees from the set of $B\cdot B_{\text{keep}}$ decision trees created by ``grow'' and evaluate the performance of ``tree-combined prediction'' under some evaluation function. Then, $B$ decision trees with the lowest evaluation function are selected.
These operations allow us to construct $B$ decision trees simultaneously and to evaluate the performance of ``tree-combined prediction'' throughout the construction process.

Our framework has similarities with RF. RF enhances the predictive performance by adopting two types of randomness to construct a variety of decision trees. The first randomness is the use of bootstrap samples in each decision tree. The second randomness is to consider a subset of the features at each split. In RF, each decision tree is generated independently and simultaneously, and the final prediction averages the predictions of each decision tree.
Constructing $B$ decision trees simultaneously and obtaining the final prediction by averaging the predictions of each decision tree is common to our framework and RF.
However, $B$ decision trees are evaluated as ``tree-combined prediction'' throughout the construction process in our framework, whereas $B$ decision trees are evaluated independently.
Therefore, we conduct an experimental comparison between our proposal and RF to show the effectiveness of evaluating the performance of ``tree-combined prediction'' throughout the construction process.

Our contributions are summarized as follows.
\begin{enumerate}
    \item We propose a new algorithmic framework for constructing $B$ decision trees that is different from bagging and boosting.
    \item In our framework, we evaluate the performance of ``tree-combined prediction'' throughout the construction process. This is based on the idea that it is natural to evaluate the performance of ``tree-combined prediction'' throughout the construction process when considering ``tree-combined prediction'' as the final prediction.
    \item We show the effectiveness of evaluating the performance of ``tree-combined prediction'' throughout the construction process by comparing our framework with RF.
\end{enumerate}

The rest of this paper is organized as follows. In Section~\ref{section2}, we describe a problem setup and define notations. In Section~\ref{section3}, we explain the proposed framework and two operations required for the framework. In Section~\ref{section4}, we confirm the effectiveness of the proposal through two experiments.

\section{Preliminaries}
\label{section2}
\subsection{Problem Setup}
In this study, we consider the regression problem of predicting an objective variable $y_{n+1}$ given training data $\mathcal{D}=\{(\bm{x}_i,y_i)\}_{i=1}^{n}$ and a new explanatory variable $\bm{x}_{n+1}$. 
For this problem, we consider using $B$ decision trees. First, we construct $B$ decision trees by the proposed framework detailed in Section~\ref{section3}. Next, we combine the predictions of each tree by averaging.
 
\subsection{Notations}
\begin{figure}[ht]
    \centering
    \includegraphics[width=0.7\hsize]{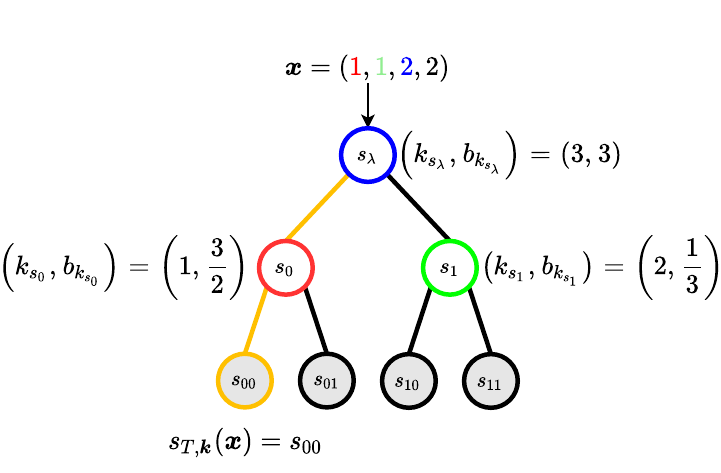}
    \caption{An example of the notations. In this figure, the parameters are the following: $D_{\text{max}}=2$, $\bm{k}=((3,3),(1,\frac{3}{2}),(2,\frac{1}{3}))$, $\mathcal{I}_T=\{s_\lambda,s_0,s_1\}$, $\mathcal{L}_T=\{s_{00},s_{01},s_{10},s_{11}\}$ (painted gray), and $s_{T,\bm{k}}(\bm{x})=s_{00}$. }
    \label{fig:function}
\end{figure}

We summarize the notations used in this paper (see also Figure~\ref{fig:function}).
\begin{dfn}
  \label{notations}
  We define the notations required for an explanation of our proposed framework.
  \begin{itemize}
    \item $p\in\mathbb{N}\cup\{0\}$ denotes the dimension of continuous explanatory variables and $q\in\mathbb{N}\cup\{0\}$ denotes the dimension of discrete explanatory variables.
    \item $T$ denotes a full binary tree whose depth is equal to or smaller than $D_{\text{max}}$.
    \item $s$ denotes a node of $T$. $\mathcal{I}_T$ and $\mathcal{L}_T$ respectively represent the set of all inner nodes and leaf nodes of $T$.
    \item $s_{\lambda}$ denotes the root node of $T$.
    \item $k_s\in\{1,2,\cdots,p+q\}$ denotes a feature index assigned to each inner node $s\in\mathcal{I}_{T}$ and $b_{k_s}\in\mathbb{R}$ denotes a threshold corresponding to $k_s$.
    $(k_s,b_{k_s})$ represents split information at $s\in\mathcal{I}_T$ and $\bm{k}=\left((k_s,b_{k_s})\right)_{s\in\mathcal{I}_T}$ represents ``split information vector.''
    \item $s_{T,\bm{k}}(\bm{x})$ denotes a leaf node $s\in\mathcal{L}_T$ which $\bm{x}$ reach.
    \item $\mathcal{Y}_s$ denotes a subset of $\{y_i\}_{i=1}^{n}$ assigned to $s$.
  \end{itemize}
\end{dfn}

\subsection{Impurity and Gain Function}
We define the impurity and the gain function used in the following section. These are the same as those used in CART and RF for regression.
\begin{dfn}
    We define the impurity for $s$ as follows:
    \begin{equation}
        i(s) \coloneqq \frac{1}{|\mathcal{Y}_s|}\sum_{y\in\mathcal{Y}_s}(y-\bar{y}_s)^2,
    \end{equation}
    where
    \begin{equation}
        \bar{y}_s \coloneqq \frac{1}{|\mathcal{Y}_s|}\sum_{y\in\mathcal{Y}_s} y.
    \end{equation}
\end{dfn}

\begin{dfn}
    \label{def:gain}
    $s_L$ and $s_R$ denote the left child node of $s$ and the right child node of $s$ respectively. We define the gain function as follows:
    \begin{equation}
        I(s) \coloneqq \frac{|\mathcal{Y}_s|}{|\mathcal{Y}_{s_\lambda}|}\left(i(s)-\frac{|\mathcal{Y}_{s_L}|}{|\mathcal{Y}_{s}|}i(s_L)-\frac{|\mathcal{Y}_{s_R}|}{|\mathcal{Y}_{s}|}i(s_R)\right).
    \end{equation}
\end{dfn}

\section{Proposed Framework}
\label{section3}
In this study, we propose a new algorithmic framework for constructing $B$ decision trees for prediction. 
When considering the prediction from $B$ decision trees, the predictions of each decision tree are combined by weighting. Therefore, it is natural to construct $B$ decision trees by evaluating the performance of ``tree-combined prediction'' throughout the construction process.  This idea is not achieved by bagging and boosting as described in Section~\ref{section1}. In contrast, we construct $B$ decision trees by evaluating the performance of ``tree-combined prediction'' throughout the construction process in our framework. 

\subsection{Basic Algorithm}
\begin{figure}[ht]
    \centering
    \includegraphics[width=0.7\hsize]{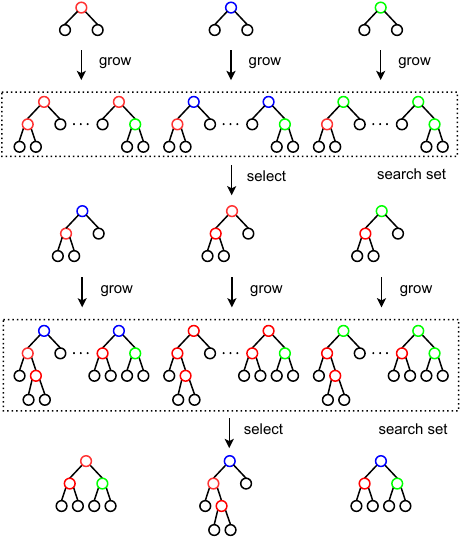}
    \caption{An example of the basic algorithm ($B=3$). In this figure, ``grow'' and ``select'' are performed twice each. We repeat these operations to construct $B$ decision trees for the final prediction.}
    \label{fig: basic}
    \end{figure}

Algorithm~\ref{alg: basic} describes the basic algorithm of our framework. An illustration of the basic algorithm is given in Figure~\ref{fig: basic}. 

\begin{algorithm}[ht]
  \caption{Basic algorithm}
  \label{alg: basic}
  \begin{algorithmic}[1]
  \Require $B$: the number of decision trees,
  $L(\cdot)$: the evaluation function, $E_{\text{min}}$: the minimum value of the evaluation function, $t_{\text{max}}$: the number of loops, $\mathcal{D}$: the training data
  \Ensure $M$: the combination of $B$ decision trees
  \State $t\gets 0$
  \State $M\gets \{(s_{\lambda},\cdot)\}$ \Comment{$\mathcal{D}$ is assigned to $s_{\lambda}$.}
  \Repeat
  \State $S_t\gets \emptyset$
  \For{$(T,\bm{k})\in M$}
  \State $S_{T,\bm{k}}\gets \bm{\mbox{grow}}\left((T,\bm{k})\right)$
  \State $S_t\gets S_t\cup S_{T,\bm{k}}$
  \EndFor
  \State $M\gets\bm{\mbox{select}}(S_t,B,L)$
  \State $\text{err}\gets L(M)$
  \State $t\gets t+1$
  \Until{$\text{err}\leq E_{\text{min}}$ or $t=t_{\text{max}}$}
    \State\textbf{return}\,\,\,$M$
    
  \end{algorithmic}
  \end{algorithm}

The basic algorithm consists of two operations named ``grow'' and ``select.'' These operations are iterated until the stopping conditions are satisfied. In ``grow,'' the input decision tree $(T,\bm{k})$ is split at one leaf node, and new decision trees are created. A set of decision trees $S_t$ is constructed by performing ``grow'' on $B$ decision trees. In ``select,'' $B$ decision trees are selected by evaluating the performance of ``tree-combined prediction'' under an evaluation function $L(\cdot)$. When the number of loops reaches $t_{\text{max}}$ or the evaluation function becomes equal to or lower than $E_{\text{min}}$, the basic algorithm stops. We introduce the details of the evaluation function $L(\cdot)$, ``grow'' and ``select'' in the next subsection and thereafter.

\subsection{Evaluation Function for a Combination Performance of Decision Trees}
\label{section3.2}
In our framework, we need to evaluate the performance of ``tree-combined prediction'' throughout the construction process. First, we define the prediction of each decision tree $(T,\bm{k})$.

\begin{dfn}
  \label{dfn: prediction}
  We define the prediction $\hat{y}_{s_{T,\bm{k}}(\bm{x})}$ corresponding to an explanatory variable $\bm{x}$ at leaf node $s$ of $(T,\bm{k})$ as follows: 
  \begin{align}
    \hat{y}_{s_{T,\bm{k}}(\bm{x})}&\coloneqq
      \displaystyle \frac{1}{|\mathcal{Y}_{s_{T,\bm{k}}(\bm{x})}|}\sum_{y\in\mathcal{Y}_{s_{T,\bm{k}}(\bm{x})}} y,\\
      &=\bar{y}_{s_{T,\bm{k}}(\bm{x})}. 
  \end{align}
\end{dfn}

The prediction defined by Definition~\ref{dfn: prediction} represents the mean of $y$ assigned to each leaf node.

Next, we define ``tree-combined prediction.'' 
\begin{dfn}
\label{dfn:tcp}
    $M$ denotes a combination of decision trees. We define ``tree-combined prediction'' as follows: 
    \begin{equation}
        \hat{y}(\bm{x})\coloneqq
    \displaystyle \frac{1}{|M|}\sum_{(T,\bm{k})\in M}\hat{y}_{s_{T,\bm{k}}(\bm{x})}.
    \end{equation}
\end{dfn}

The prediction defined in Definition~\ref{dfn:tcp} is the average for the predictions of each decision tree.

Finally, we define the evaluation function.
\begin{dfn}
\label{def:L}
  We define the evaluation function for the combination of decision trees $M$ as follows: 
\begin{align}
  L(M)&\coloneqq
    \displaystyle \frac{1}{n}\sum_{i=1}^{n}(y_i- \hat{y}(\bm{x}_i))^2.
\end{align}
\end{dfn}

This evaluation function calculates the mean squared error (MSE) between $y$ of the training data and ``tree-combined prediction.''

\subsection{Operation ``Grow''}
\begin{figure}[ht]
  \centering
  \includegraphics[width=0.7\hsize]{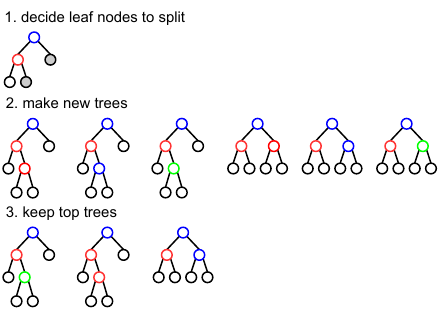}
  \caption{An example of ``grow.'' In this figure, the parameters are the following: $m_{\text{leaf}}=2$, $B_{\text{keep}}=3$. Therefore, we determine the top two leaf nodes that have the largest $h(s)$, make new trees, and keep the top three trees that have the largest gain function.
  }
  \label{fig: grow}
  \end{figure}

``Grow'' is an operation that creates new $B_{\text{keep}}$ decision trees from the input decision tree $(T,\bm{k})$. We perform ``grow'' for $B$ decision trees at step $t$, so a set of $B\cdot B_{\text{keep}}$ decision trees $S_t$ is obtained. This set (the search set) is used for ``select'' to find new $B$ decision trees. We summarize the ``grow'' in Algorithm~\ref{alg: grow} and Figure~\ref{fig: grow}. In addition, we describe the functions used in the Algorithm~\ref{alg: grow} as follows.
\begin{itemize}
    \item $\text{GetSplitLeaves}(T,l,h)$: this function determines $l$ leaf nodes of $T$ with the largest heuristic $h(s)$ from $\mathcal{L}_T$. These selected leaf nodes will be split in the following steps.
    \item $\text{GenerateSplitRules}(s)$: 
    this function generates candidates for split information at $s$. Therefore, this function generates a set of $(k_s,b_{k_s})$.
    $\mathcal{X}_s^{k}$ denotes the subset of $\{x_{ik}\}_{i=1}^{n}$ assigned to $s$.
    Then, $\lfloor\log_2|\mathcal{Y}_s|+1\rfloor$ thresholds are generated from the interval 
$\displaystyle\left[\min_{x\in\mathcal{X}_s^{k_s}}x,\max_{x\in\mathcal{X}_s^{k_s}}x\right]$ uniformly for each $k$.
    \item $\text{Split}((T,\bm{k}),s,(k_s,b_{k_s}))$: 
    this function splits leaf node $s$ of $(T,\bm{k})$ by $(k_s,b_{k_s})$ and generate a new tree. Then, the training data corresponding to the split is assigned to new child nodes $s_L$ and $ s_R$.
    \item $\text{GetTopTrees}(S,l)$: this function selects $l$ decision trees from $S$ with the largest gain function.
    
  \end{itemize}
  
\begin{algorithm}[ht]
\caption{Operation ``grow''}
\label{alg: grow}
\begin{algorithmic}[1]
\Require $(T,\bm{k})$: the decision tree, $h(\cdot)$: the heuristic of leaf nodes, $m_{\text{leaf}}$: the number of leaf nodes to split, 
$B_{\text{keep}}$: the number of decision trees to keep
\Ensure $S_{T,\bm{k}}$: the set of $B_{\text{keep}}$ decision trees.
\State $S_{T,\bm{k}}\gets \emptyset$
\State $\mathcal{L}\gets \text{GetSplitLeaves}(T,m_{\text{leaf}},h)$
\For{$s\in\mathcal{L}$}
\State $\mathcal{K}\gets \text{GenerateSplitRules}(s)$ 
\For{$(k_s,b_{k_s})\in \mathcal{K}$}
\State $(T_{\text{new}},\bm{k}_{\text{new}})\gets \text{Split}((T,\bm{k}),s,(k_s,b_{k_s}))$
\State $S_{T,\bm{k}}\gets S_{T,\bm{k}}\cup \{(T_{\text{new}},\bm{k}_{\text{new}})\}$ 
\EndFor
\EndFor
\State $S_{T,\bm{k}}\gets \text{GetTopTrees}(S_{T,\bm{k}},B_{\text{keep}})$
\State\textbf{return}\,\,\,$S_{T,\bm{k}}$
\end{algorithmic}
\end{algorithm}

``Grow'' can be separated into the following three operations.
\begin{itemize}
    \item[(A)] Candidates of the leaf nodes to split are determined.
    \item[(B)] New decision trees are created by splitting.
    \item[(C)] $B_{\text{keep}}$ decision trees are kept.
\end{itemize}

In (A), we assign a heuristic $h(s)$ to each leaf node and select the top $m_{\text{leaf}}$ leaf nodes with the largest $h(s)$. In this study, we use the inverse of depth of $s$ for $h(s)$. This leads to a search similar to the breadth-first search.

In (B), we generate the candidates of split information $(k_s,b_{k_s})$ at each leaf node selected in (A). In this study, we generate $\lfloor\log_2|\mathcal{Y}_s|+1\rfloor$ thresholds $b_{k_s}$ from the interval $\displaystyle\left[\min_{x\in\mathcal{X}_s^{k_s}}x,\max_{x\in\mathcal{X}_s^{k_s}}x\right]$ uniformly for each $k_s\in\mathcal\{1,2,\cdots,p+q\}$. Therefore, up to $m_{\text{leaf}}\lfloor\log_2|\mathcal{Y}_s|+1\rfloor(p+q)$ split information are generated.
We use the candidates of split information to split each leaf node and generate new decision trees. The training data corresponding to the split are assigned to the newly created child nodes $s_L$ and $s_R$. Therefore, the following training data are assigned to each child node.
\begin{align}
    \mathcal{D}_{s_L} &= \left\{(\bm{x},y) \in \mathcal{D}_s \mid x_{k_s} < b_{k_s}  \right\},\\
    \mathcal{D}_{s_R} &= \left\{(\bm{x},y) \in \mathcal{D}_s \mid x_{k_s} \geq b_{k_s}  \right\},
\end{align}
where $\mathcal{D}_s$ denotes the training data assigned to $s$.

In (C), we select the top $B_{\text{keep}}$ trees with the largest gain function. This means that only decision trees with efficient splitting are kept. The gain function is defined in Definition~\ref{def:gain}.

\subsection{Operation ``Select''}
``Select'' is an operation that selects $B$ decision trees from a search set $S_t$. $B$ decision trees are evaluated using Definition~\ref{def:L}. In other words, $B$ decision trees are selected based on evaluating the performance of ``tree-combined prediction.'' There are $_{|S_t|}C_{B}$ possible combinations when selecting $B$ decision trees from $S_t$. Therefore, we need to evaluate the performance for $_{|S_t|}C_{B}$ combinations of $B$ decision trees. This is not realistic in terms of computational order. 
We show the following ideas to reduce the number of combinations that evaluated the performance. 

\begin{itemize}
  \item[(X)] Reduction for the number of combinations
  \item[(Y)] Reduction for the elements of the search set
\end{itemize} 

\begin{figure}[ht]
  \centering
  \includegraphics[width=0.7\hsize]{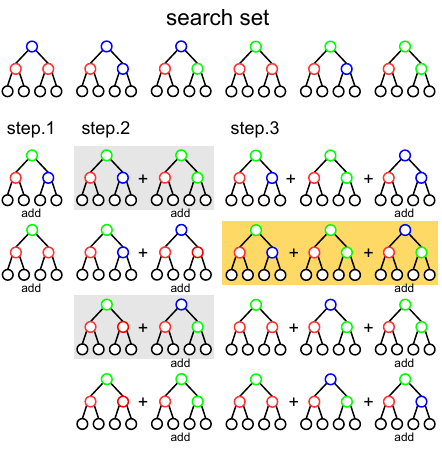}
  \caption{An example of the greedy search. 
  In this figure, the parameters are the following: $B=3$, $C=2$. First, we evaluate the performance of the prediction by a single decision tree with the evaluation function and keep the top two decision trees with the lowest evaluation function. Next, we consider ``tree-combined prediction,'' combining the first tree with the newly added tree. Then, we evaluate the performance of each ``tree-combined prediction'' and keep the top two combinations with the lowest evaluation function. In this example, we have four combinations because we keep the top two trees first and each tree is combined with the new tree, but we keep only the two with the lowest evaluation function (painted gray). Finally, we consider ``tree-combined prediction,'' combining the selected two trees with the newly added tree. Then, we evaluate the performance of each ``tree-combined prediction'' and keep the combination with the lowest evaluation function (painted yellow).}
  \label{fig: greedy}
  \end{figure}

(X) is the idea that we evaluate the performance of ``tree-combined prediction'' for some combinations, not all combinations, based on a search algorithm. In this study, we introduce a greedy search (GS) as the search algorithm. We summarize GS in Algorithm~\ref{alg: greedy} and Figure~\ref{fig: greedy}. In addition, we describe the functions used in the Algorithm~\ref{alg: greedy} as follows.
\begin{itemize}
  \item $\text{RandomPick}(S,l)$: 
  this function picks $l$ elements from $S$ without replacement.
  \item $\text{GetTopCombs}(\mathcal{M},l,L)$: 
  this function selects $l$ combinations with the lowest $L(M)$ from $\mathcal{M}$ where $\mathcal{M}$ is a set of combinations.
\end{itemize}

\begin{algorithm}[ht]
\caption{Greedy search}
\label{alg: greedy}
\begin{algorithmic}[1]
\Require $S$: the search set, $B$ the number of decision trees, $\gamma$: the rate, $C$: the number of combinations, 
$L(\cdot)$: the evaluation function
\Ensure $M$: the combination of $B$ decision trees. 
\Function{\textnormal{GreedySearch}}{$M,S,C,L,\gamma$}
\State $\mathcal{M}_{C}\gets \emptyset$
\State $S_{\text{pick}}\gets \text{RandomPick}(S,\gamma|S|)$
\For{$(T,\bm{k})\in S_{\text{pick}}$}
\State $M_{\text{new}}\gets M\cup \{(T,\bm{k})\}$
\If{$|\mathcal{M}_{C}|<C$}
\State $\mathcal{M}_{C}\gets \mathcal{M}_{C}\cup \{M_{\text{new}}\}$
\Else
\If{$\displaystyle L(M_{\text{new}})<\max_{M^*\in \mathcal{M}_{C}}L(M^*)$}
\State $\mathcal{M}_{C}\gets \mathcal{M}_{C} \setminus \{M^*\}$
\State $\mathcal{M}_{C}\gets \mathcal{M}_{C} \cup \{M_{\text{new}}\}$
\EndIf
\EndIf
\EndFor
\State \textbf{return}\,\,\,$\mathcal{M}_{C}$
\EndFunction
\State $\mathcal{M}_0\gets\{\emptyset\}$
\For{$b=1,2,\cdots, B$}
\State $\mathcal{M}_b=\emptyset$
\For{$M\in \mathcal{M}_{b-1}$}
\State $\mathcal{M}^*\gets \text{GreedySearch}(M,S,C,L,\gamma)$
\State $\mathcal{M}_b\gets \mathcal{M}_b\cup \mathcal{M}^*$
\EndFor
\State $\mathcal{M}_b\gets \text{GetTopCombs}(\mathcal{M}_b,C,L)$
\EndFor
\State\textbf{return}\,\,\,$\displaystyle \min_{M\in\mathcal{M}_B}L(M)$
\end{algorithmic}
\end{algorithm}

In GS, decision trees are added one by one, and a combination of $B$ decision trees is created. When adding a new decision tree to create a combination of $b$ decision trees, we consider ``tree-combined prediction'' with the already chosen $b-1$ decision trees and the new decision tree. We evaluate new ``tree-combined prediction'' based on $L(M)$ defined in Definition~\ref{def:L} and keep the top $C$ combinations with the lowest $L(M)$. We iterate this operation and get $B$ decision trees. 

The idea of (Y) is to reduce the number of combinations by constraining the elements of the search set.
We introduce the following two ideas as examples.
\begin{itemize}
    \item[(Y-1)] Random picking
    \item[(Y-2)] Splitting the search set into blocks 
\end{itemize}

(Y-1) is the idea that we randomly pick decision trees from the search set $S_t$ and make the new search set $S_{\text{pick}}$ where $|S_{\text{pick}}|$ is smaller than $|S_t|$.
This idea is built into Algorithm \ref{alg: greedy} where $\gamma$ is the parameter that determines the size of $S_{\text{pick}}$. If we set $\gamma=1$, $S_{\text{pick}}=S_t$.

\begin{figure}[ht]
  \centering
  \includegraphics[width=0.9\hsize]{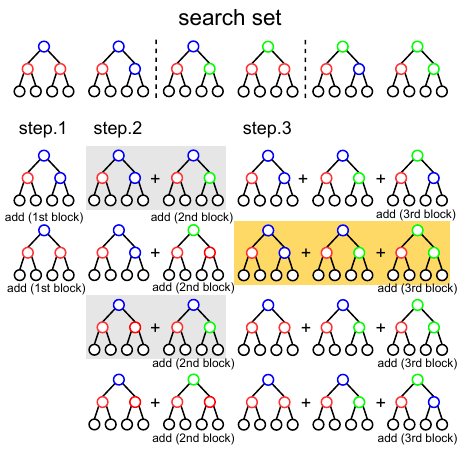}
  \caption{An example of the blocked greedy search. 
  In this figure, the parameters are the following: $B=3$, $C=2$. The difference for the greedy search is the search set used for adding a new tree. When adding a new tree, it is necessary to select a tree from the corresponding block.}
  \label{fig: block_greedy}
  \end{figure}

(Y-2) is the idea that we divide the search set $S_t$ into multiple blocks and conduct GS for each block.
We summarize the algorithm that adapts (Y-2) for GS in Algorithm \ref{alg: block-greedy} and Figure~\ref{fig: block_greedy}.
We call this algorithm a ``blocked greedy search'' (BGS).
Here, $\text{SplitSearchSet}(S,B)$ denotes the operation of splitting $S$ into $B$ blocks in the order of the decision trees. 
The search set $S_t$ consists of $B_{\text{keep}}$ decision trees added $B$ times. $B_{\text{keep}}$ decision trees were the same decision tree before adapting ``grow,'' so similar decision trees are lined up close. Therefore, when the search set is split in order, each block contains similar trees.

\begin{algorithm}[ht]
\caption{Blocked greedy search}
\label{alg: block-greedy}
\begin{algorithmic}[1]
\Require $S$: the search set, $B$ the number of decision trees, $\gamma$: the rate, $C$: the number of combinations, 
$L(\cdot)$: the evaluation function
\Ensure $M$: the combination of $B$ decision trees. 
\State $S_1,S_2,\cdots,S_B\gets \text{SplitSearchSet}(S,B)$
\State $\mathcal{M}_0\gets\{\emptyset\}$
\For{$b=1,2,\cdots, B$}
\State $\mathcal{M}_b\gets\emptyset$
\For{$M\in \mathcal{M}_{b-1}$}
\State $\mathcal{M}^*\gets\text{GreedySearch}(M,S_b,C,L,\gamma)$
\State $\mathcal{M}_b\gets \mathcal{M}_b\cup \mathcal{M}^*$
\EndFor
\State $\mathcal{M}_b\gets \text{GetTopCombs}(\mathcal{M}_b,C,L)$
\EndFor
\State\textbf{return}\,\,\,$\displaystyle \min_{M\in\mathcal{M}_B}L(M)$
\end{algorithmic}
\end{algorithm}

\section{Experiments}
\label{section4}
In this section, we conduct two experiments to confirm the performance of the proposed framework. Both our framework and bagging construct $B$ decision trees simultaneously and predict by averaging the predictions of each decision tree. However, they differ in the target of evaluation throughout the construction process. In our framework, the performance of ``tree-combined prediction'' is evaluated throughout the construction process, but each decision tree is evaluated independently throughout the construction process in bagging. Therefore, a comparison of our framework and bagging is appropriate to see the effectiveness of evaluating the performance of ``tree-combined prediction'' throughout the construction process is effective. Since the most famous method of bagging is RF, we compare our framework and RF in this study. Our framework allows for two different algorithms depending on whether GS or BGS is used in ``select'' for the basic algorithm. Therefore, we compare GS (use GS for ``select''), BGS (use BGS for ``select''), and RF. In the two experiments, we use the MSE which is a common metric for evaluating experiments on regression. We also use the MSE for the evaluation function defined in Definition~\ref{def:L}, so it is the appropriate metric.

\subsection{Parameter Settings}
We use RF implemented in scikit-learn 1.3.2~\cite{scikit-learn}. The parameters of RF are set to default (e.g. max depth $D_{\text{max}}=\text{None}$, the number of decision trees $B=100$). For our framework, we set the common parameters for  $D_{\text{max}}=\text{None}$ and $B=100$. Other parameter settings are as follows: $C=5$, $\gamma=1$, $m_{\text{leaf}}=5$, $B_{\text{keep}}=5$, $t_{\text{max}}=500$, $E_{\text{min}}=10^{-2}$. These are our default parameters determined in terms of computational complexity. In the two experiments, the parameters of each algorithm are not tuned because we confirm the average performance. It is probably better to set the parameters other than $E_{\text{min}}$ larger in our framework.

\subsection{MSE for Synthetic Data}
\textbf{Purpose.}
We compare the proposed methods with RF using synthetic data generated from complicated functions and confirm the effectiveness of evaluating the performance of ``tree-combined prediction'' throughout the construction process.
We also confirm the robustness of the proposed methods, since we can add noise to synthetic data.

\noindent\textbf{Data.}
We use the following data used in~\cite{Bagging}: Friedman 1, Friedman 2, and Friedman 3. 
We describe the generation of the data as follows.

\noindent Friedman 1: 
The ten continuous explanatory variables are generated independently and uniformly from the interval $[0,1]$. The objective variable is generated from the following equation. 
\begin{equation}
  y=10\sin (\pi x_1x_2) +20(x_3-5)^2+10x_4+5x_5 + \alpha \varepsilon.
\end{equation}
    
\noindent Friedman 2: The four continuous explanatory variables are generated uniformly from $0\leq x_1\leq 100$, $\displaystyle 20\leq\frac{x_2}{2\pi}\leq 280$, $0\leq x_3\leq 1$, and $1\leq x_4 \leq 11$, respectively. The objective variable is generated from the following equation. 
\begin{equation}
  y=\left(x_1^2+\left(x_2x_3-\frac{1}{x_2x_4}\right)^2\right)^{\frac{1}{2}}+ \alpha \varepsilon.
\end{equation}

\noindent Friedman 3: The four explanatory variables are generated uniformly from $0\leq x_1\leq 100$, $\displaystyle 20\leq\frac{x_2}{2\pi}\leq 280$, $0\leq x_3\leq 1$, and $1\leq x_4 \leq 11$, respectively. The objective variable is generated from the following equation. 
\begin{equation}
  y=\tan^{-1}\left(\frac{x_2x_3-\frac{1}{x_2x_4}}{x_1}\right)+ \alpha \varepsilon.
\end{equation}

$\varepsilon$ is the noise generated from the standard normal distribution $\mathcal{N}(0,1)$ and $\alpha$ is the parameter which determines the level of noise.
Each data can be generated by using the function implemented in scikit-learn 1.3.2~\cite{scikit-learn}. We summarize the details of each generated data in Table~\ref{tab:friedman}.
    \begin{table}[ht]
    \centering
    \caption{A summary of the generated data}
    \vspace*{0.3cm}
    \label{tab:friedman}
    \begin{tabular}{c|cccc}
    \toprule
        Data & $n$ & $p$ & $q$ & $\alpha$\\
    \midrule
        Friedman 1& 1000& 10& 0&0\\
        Friedman 2 & 1000& 4& 0&0\\
        Friedman 3& 1000& 4& 0&0\\
        \midrule
        Friedman 1*& 1000& 10& 0&1\\
        Friedman 2*& 1000& 4& 0&1\\
        Friedman 3*& 1000& 4& 0&1\\
    \bottomrule
    \end{tabular}
\end{table}

After generating the data, we standardize the continuous explanatory variables and the objective variable for all datasets to align the scales. 


\noindent\textbf{Procedures.}
We perform 5-fold cross-validation three times and output the average of the MSE for each method. 

\noindent\textbf{Result.}
Table~\ref{tab:mse-friedman} shows the MSE of each synthetic data for each method. From Table~\ref{tab:mse-friedman}, we can see that the MSE of RF is larger than that of the proposed methods for Friedman 1 and Friedman 3. For Friedman 2, the MSE of RF is the lowest. 
This is because the stopping condition of the proposed methods is $E_{\text{min}}=10^{-2}$ and training is stopped in the middle of the construction process.
If noise is included in each data, evaluating the performance of ``tree-combined prediction'' throughout the construction process could lead to bad results. However, the results in Table~\ref{tab:mse-friedman} show that BFS has a smaller MSE than RF, even in the presence of noise. 
We can think that this is due to the difference between the search set of GS and that of BGS.
In GS, all $B$ decision trees are selected from the same search set, whereas in BGS, $B$ decision trees are selected from different search sets.
Therefore, if the evaluation function is lower in GS, the combination of similar decision trees may be selected. In BGS, the combination of various decision trees is chosen because similar decision trees are contained in the same block. We can think of this diversity as affecting robustness.

\begin{table}[ht]
    \centering
    \caption{The MSE of each method for the synthetic data}
    \vspace*{0.3cm}
    \label{tab:mse-friedman}
    \begin{tabular}{c|ccc}
    \toprule
        method & GS & BGS & RF \\ \hline
        Friedman 1 & 0.1108 & \textbf{0.1010} & 0.1225 \\ 
        Friedman 2 & 0.0119 & 0.0116 & \textbf{0.0032} \\ 
        Friedman 3 & 0.0441 & \textbf{0.0392} & 0.0482 \\ \midrule
        Friedman 1* & 0.1714 & \textbf{0.1519}& 0.1740 \\ 
        Friedman 2* & 0.0110 & 0.0111 & \textbf{0.0026} \\ 
        Friedman 3* & 1.1941 & \textbf{0.781} & 1.0005 \\ 
    \bottomrule
    \end{tabular}
\end{table}

\subsection{MSE for Famous Benchmark Datasets}
\textbf{Purpose.}
We compare the proposed methods with RF using well-known benchmark datasets and confirm the effectiveness of evaluating the performance of ``tree-combined prediction'' throughout the construction process.

\noindent\textbf{Datasets.}
We use the dataset used in the experiment of RF~\cite{breiman2001random}.
We summarize the details of each dataset in Table~\ref{tab:dataset}. Abalone and Servo are from the UCI repository~\cite{UCI}, Boston is from \cite{boston}, and Ozone is from \cite{ozone}.
    \begin{table}[ht]
    \centering
    \caption{A summary of the datasets}
    \vspace*{0.3cm}
    \label{tab:dataset}
    \begin{tabular}{c|ccc}
    \toprule
        Dataset & $n$ & $p$ & $q$\\
    \midrule
        Abalone & 4177& 7& 1\\
        Boston & 506& 12& 1\\
        Ozone & 330& 8& 0\\
        Servo & 167& 2& 2\\
    \bottomrule
    \end{tabular}
\end{table}

We preprocess all datasets in the following steps.
First, we remove the data containing missing values. Next, we standardize the continuous explanatory variables and the objective variable for all datasets to align the scales. 
Finally, we use label encoding for discrete explanatory variables of ordinal scales and one-hot encoding for discrete explanatory variables of nominal scales.


\noindent\textbf{Procedures.}
We perform 5-fold cross-validation three times and output the average of the MSE. 

\noindent\textbf{Result.}
Table~\ref{tab:mse} shows the MSE of each dataset for each method. From this table, 
we can see that the MSE of BGS is lower than that of other algorithms for each dataset. 
However, the MSE of GS is larger than that of RF except for the Boston dataset. 
The experiment of benchmark datasets also shows a similar result in the experiment that noise is added to the synthetic data. 
Therefore, it is considered important to evaluate the performance of ``tree-combined prediction'' for the combination of various decision trees throughout the construction process, even when we consider the prediction on benchmark datasets.

\begin{table}[ht]
    \centering
    \caption{The MSE of each method for the benchmark datasets}
    \vspace*{0.3cm}
    \label{tab:mse}
    \begin{tabular}{c|ccc}
    \toprule
        dataset & GS & BGS & RF \\ \midrule
        Abalone & 0.4979 & \textbf{0.4551} & 0.4622 \\
        Boston & 0.1360 & \textbf{0.1278} & 0.1497 \\
        Ozone & 0.3138 & \textbf{0.2861} & 0.2881 \\
        Servo & 0.1160 & \textbf{0.0946} & 0.1091 \\ \bottomrule
    \end{tabular}
\end{table}

\section{Conclusion and Future Work}
In this paper, we proposed the algorithmic framework for constructing $B$ decision trees used for the prediction. 
Bagging and boosting, on which many decision tree algorithms are based, do not evaluate the performance of ``tree-combined prediction'' throughout the construction process. 
On the other hand, We construct $B$ decision trees simultaneously and evaluate the performance of ``tree-combined prediction'' throughout the construction process.
We confirmed the effectiveness of evaluating the performance of ``tree-combined prediction'' throughout the construction process by conducting experiments.

In this paper, we proposed the general framework for constructing $B$ decision trees. Operations ``grow'' and ``select'' introduced in this paper are only some examples, and various algorithms can be considered based on this framework.
In addition, our framework can be applied to classification as well, although we focused on regression in this paper. In that case, the prediction and the evaluation function defined in Section~\ref{section3.2} should be adapted for classification.

\bibliography{hoge}

\begin{thebibliography}{10}

\bibitem{CART}
L.~Breiman, J.~Friedman, C.J. Stone, and R.A. Olshen.
\newblock {\em Classification and Regression Trees}.
\newblock Taylor \& Francis, 1984.

\bibitem{ozone}
L.~Breiman and Jerome~H. Friedman.
\newblock Estimating optimal transformations for multiple regression and
  correlation.
\newblock {\em Journal of the American Statistical Association}, 80:580--598,
  1985.

\bibitem{Bagging}
Leo Breiman.
\newblock Bagging predictors.
\newblock {\em Machine learning}, 24(2):123--140, 1996.

\bibitem{breiman2001random}
Leo Breiman.
\newblock Random forests.
\newblock {\em Machine learning}, 45(1):5--32, 2001.

\bibitem{XGBoost}
Tianqi Chen and Carlos Guestrin.
\newblock {XGBoost: A Scalable Tree Boosting System}.
\newblock In {\em Proceedings of the 22nd ACM SIGKDD International Conference
  on Knowledge Discovery and Data Mining}, KDD '16, pages 785--794, New York,
  NY, USA, 2016. Association for Computing Machinery.

\bibitem{UCI}
Dheeru Dua and Casey Graff.
\newblock {UCI} machine learning repository, 2017.

\bibitem{Boosting}
Yoav Freund and Robert~E Schapire.
\newblock A decision-theoretic generalization of on-line learning and an
  application to boosting.
\newblock {\em Journal of Computer and System Sciences}, 55(1):119--139, 1997.

\bibitem{GBDT}
Jerome~H. Friedman.
\newblock {Greedy function approximation: A gradient boosting machine.}
\newblock {\em The Annals of Statistics}, 29(5):1189 -- 1232, 2001.

\bibitem{ET}
Pierre Geurts, Damien Ernst, and Louis Wehenkel.
\newblock {Extremely Randomized Trees}.
\newblock {\em {Machine Learning}}, 36:3--42, 2006.

\bibitem{boston}
David Harrison and Daniel~L. Rubinfeld.
\newblock Hedonic housing prices and the demand for clean air.
\newblock {\em Journal of Environmental Economics and Management},
  5(1):81--102, 1978.

\bibitem{LightGBM}
Guolin Ke, Qi~Meng, Thomas Finley, Taifeng Wang, Wei Chen, Weidong Ma, Qiwei
  Ye, and Tie-Yan Liu.
\newblock {LightGBM: A Highly Efficient Gradient Boosting Decision Tree}.
\newblock In I.~Guyon, U.~Von Luxburg, S.~Bengio, H.~Wallach, R.~Fergus,
  S.~Vishwanathan, and R.~Garnett, editors, {\em Advances in Neural Information
  Processing Systems}, volume~30, pages 3146--3154. Curran Associates, Inc.,
  2017.

\bibitem{scikit-learn}
F.~Pedregosa, G.~Varoquaux, A.~Gramfort, V.~Michel, B.~Thirion, O.~Grisel,
  M.~Blondel, P.~Prettenhofer, R.~Weiss, V.~Dubourg, J.~Vanderplas, A.~Passos,
  D.~Cournapeau, M.~Brucher, M.~Perrot, and E.~Duchesnay.
\newblock Scikit-learn: Machine learning in {P}ython.
\newblock {\em Journal of Machine Learning Research}, 12:2825--2830, 2011.

\bibitem{ID3}
J.~Ross Quinlan.
\newblock Induction of decision trees.
\newblock {\em Machine Learning}, 1:81--106, 1986.

\bibitem{C4.5}
J.~Ross Quinlan.
\newblock {\em C4.5: Programs for Machine Learning}.
\newblock Morgan Kaufmann Publishers Inc., San Francisco, CA, USA, 1993.

\end{thebibliography}
\bibliographystyle{plain}

\end{document}